\def\year{2017}\relax
\documentclass[letterpaper]{article}
\usepackage{aaai17}
\usepackage{times}
\usepackage{helvet}
\usepackage{courier}

\usepackage{graphicx} 
\usepackage{algorithm}
\usepackage{algorithmic}
\usepackage{amsfonts}
\usepackage{amssymb}
\usepackage{amsmath}
\usepackage{amsthm}
\usepackage{xfrac}

\theoremstyle{definition}

\newtheorem{lemma}{Lemma}[]
\newtheorem{remark}{Remark}[]

\begin{document}
%
\title{Local Discriminant Hyperalignment for multi-subject fMRI data alignment}
\author{Muhammad Yousefnezhad, Daoqiang Zhang\\
College of Computer Science and Technology, \\Nanjing University of Aeronautics and Astronautics, Nanjing, China.\\
\{myousefnezhad, dqzhang\}@nuaa.edu.cn}
\maketitle

\begin{abstract}
Multivariate Pattern (MVP) classification can map different cognitive states to the brain tasks. One of the main challenges in MVP analysis is validating the generated results across subjects. However, analyzing multi-subject fMRI data requires accurate functional alignments between neuronal activities of different subjects, which can rapidly increase the performance and robustness of the final results. Hyperalignment (HA) is one of the most effective functional alignment methods, which can be mathematically formulated by the Canonical Correlation Analysis (CCA) methods. Since HA mostly uses the unsupervised CCA techniques, its solution may not be optimized for MVP analysis. By incorporating the idea of Local Discriminant Analysis (LDA) into CCA, this paper proposes Local Discriminant Hyperalignment (LDHA) as a novel supervised HA method, which can provide better functional alignment for MVP analysis. Indeed, the locality is defined based on the stimuli categories in the train-set, where the correlation between all stimuli in the same category will be maximized and the correlation between distinct categories of stimuli approaches to near zero. Experimental studies on multi-subject MVP analysis confirm that the LDHA method achieves superior performance to other state-of-the-art HA algorithms.
\end{abstract}

\section{Introduction}
As an imaging technology, functional Magnetic Resonance Imaging (fMRI) measures neural activity by employing the Blood-Oxygen-Level-Dependent (BOLD) contrast as a proxy for neural activation. The main idea is utilizing these measurements of neural activities to shed light on cognitive processes. Indeed, fMRI enables us to ask what information is represented in a region of the human brain and how that information is encoded, instead of asking what is a region’s function \cite{haxby14}. Multivariate Pattern (MVP) classification is one of the main techniques in fMRI analysis, which can extract and decode brain patterns by applying the classification methods  \cite{haxby11,chen15,oswal16}. In fact, it can predict patterns of neural activities associated with different cognitive states \cite{mohr15,chen15,figueiredo16} and also can define decision surfaces to distinguish different stimuli for decoding the brain and understanding how it works \cite{haxby11,haxby14}. MVP analysis can be used to find novel treatments for mental diseases or even to create a new generation of the user interface. 
		
One of the main challenges in fMRI studies, especially MVP analysis, is using multi-subject datasets. On the one hand, the multi-subject analysis is critical to figure out the generality and validity of the generated results across subjects. On the other hand, analyzing multi-subject fMRI data requires accurate functional and anatomical alignments between neuronal activities of different subjects in order to increase the performance of the final results \cite{haxby11,chen14}. Indeed, the fMRI datasets must be aligned across subjects in multi-subject studies in order to take between-subject variability into account. As mentioned before, there are two main alignment approaches, i.e. anatomical alignment and functional alignment, which can work in unison. The anatomical alignment is the most common method for aligning fMRI images based on anatomical features by employing structural MRI images, e.g. Talairach alignment \cite{talairach88}. However, this method generated limited accuracy since the size, shape and anatomical location of functional loci differ across subjects \cite{watson93,rademacher93}. Indeed, anatomical alignment is just used in many fMRI studies as a preprocessing step. By contrast, functional alignment seeks to directly align the brain neural responses across subjects. 
		
Hyperalignment (HA) \cite{haxby11} is one of the most famous methods for functional alignment. HA can be mathematically formulated by Canonical Correlation Analysis (CCA). As a result, HA for multi-subject fMRI studies can be defined as a multiple-set CCA \cite{xu12,lorbert12,chen14}. Since the unsupervised CCA techniques are employed for solving HA problems, the solution may not be optimized for MVP analysis. In other words, CCA just finds a set of mappings to maximize the correlation between same time-points of functional activities \emph{(in voxel-level)} for all subjects, while it must maximize the correlation between homogeneous stimuli (from the same category) \emph{and also remove the correlation between different categories of stimuli}. Indeed, this is a common problem in Machine Learning. For instance, Linear Discriminant Analysis (LDA) is mostly used rather than Principal Component Analysis (PCA) in the classification analysis, where LDA uses the supervision information such as class labels or similarity between samples for improving the performance of classification methods. 
		
As the main contribution of this paper, we introduce Local Discriminant Hyperalignment (LDHA) method, which incorporates the idea of Local Discriminate Analysis (LDA) into CCA \cite{peng10} in order to improve the performance of the hyperalignment solution. In brief, the idea of locality is defined based on the stimuli categories (class labels) in the train-set, where the proposed method firstly generates two sets for each category of stimuli, i.e. the set of nearest homogeneous stimuli as within-class neighborhoods and the set of stimuli from distinct categories as between-class neighborhoods. Then, these two sets are used to provide a better HA solution, where the correlation between the within-class neighborhoods is maximized, and also the correlation among between-class neighborhoods approaches to near zero.
		
The rest of this paper is organized as follows: In Section 2, this study briefly reviews some related works. Then, it introduces the proposed method in Section 3. Experimental results are reported in Section 4; and finally, this paper presents conclusion and pointed out some future works in Section 5.
\section{Related Works}
There are several studies, which used functional and anatomical features for alignment. Sabuncu et al. \shortcite{sabuncu10} employed cortical warping for maximizing the inter-subject correlation between functional responses across subjects. Conroy et al. \shortcite{conroy09} also developed a method to maximize the alignment of intra-subject patterns of cortical functional connectivity by using a cortical warping. 
		
Hyperalignment (HA) is proposed by Haxby et al. \shortcite{haxby11}, which is an `anatomy free' alignment method based on functional features. HA utilized the Procrustean transformation \cite{schonemann66} to map each the functional responses of each subject into a common high-dimensional model (template). The performance of MVP analysis by using the Hyperalignment is rapidly increased in comparison with the methods that just use the anatomical alignment \cite{haxby11,haxby14}.
		
Lorbert et al. \shortcite{lorbert12} developed Kernel Hyperalignment (KHA) to conduct nonlinear hyperalignment in an embedding space. Xu et al. \shortcite{xu12} introduced the regularized Hyperalignment, which makes connections to Canonical Correlation Analysis (CCA) \cite{gower04}. Dmochowski et al. \shortcite{dmochowski12} applied correlated component analysis to maximize inter-subject correlation by aggregating the subjects’ data into an individual matrix. Sui et al. \shortcite{sui11,sui13} proposed a `multimodal CCA + joint independent component analysis (ICA)' on multimodal data to identify the unique and shared variance associated with each imaging modality. Chen et al. \shortcite{chen14} examined a two-phase joint SVD-Hyperalignment algorithm, where a joint Singular Value Decomposition (SVD) is employed to provide dimensionality reduction, then HA aligns the subjects' responses in the lower dimensional feature space. 
		
Micheal et al. \shortcite{michael15} developed the GICA, IVA algorithms, which do not assume time-synchronized stimulus. Hence, they concatenate data along the time dimension (implying spatial consistency) and learn spatial independent components. Recently, Guntupalli et al. \shortcite{guntupalli16} proposed a linear model of shared representational spaces in human cortex. This model can capture fine-scale distinctions among population responses with response-tuning basis functions that are common across brains and models cortical patterns of neural responses with individual-specific topographic basis functions.
\section{The Proposed Method}
This paper proposes a supervised version of hyperalignment method for applying the MVP classification. The procedure is so simple: such as all classification problems, there are two sets of data points, i.e. train-set and test-set. The train-set is used for generating the classification model. Then, the generated model is evaluated on the test-set. Since functional activities in different brains are originally unaligned in comparison with each other, the previous HA methods just used the data points in the train-set to generate a template for functional aligning in both the train and test sets before the MVP analysis. By contrast, our proposed method uses data points as well as \emph{class labels} in the train-set for generating the HA template (which will be denoted by $\mathbf{G}$). In train-set, the preprocessed fMRI time series collected for $S$ subjects can be defined by $\mathbf{X}^{(i)}=\Big\{\mathbf{x}^{(i)}_{mn}\Big\}\in \mathbb{R}^{T \times V}\text{, } i=1\text{:}S, m=1\text{:}T, n=1\text{:}V$, where $T$ denotes the number of time points in unites of TRs (Time of Repetition), $V$ is the number of voxels, and $\mathbf{x}^{(i)}_{mn} \in \mathbb{R}$ denotes the functional activity for the $i-th$ subject in the $m-th$ time point and the $n-th$ voxel. In test-set, we have similar notations where the data points are defined by $\mathbf{\bar{X}}^{(i)}=\Big\{\mathbf{\bar{x}}^{(i)}_{mn}\Big\}\in \mathbb{R}^{T \times V}\text{, } i=1\text{:}\bar{S}, m=1\text{:}T, n=1\text{:}V$. Here, $\bar{S}$ is the number of subjects in the test-set. In addition, the class labels in the train-set is denoted by $\mathbf{Y}=\Big\{\mathbf{y}_m \Big\}\in\mathbb{N}^{T}\text{, } m=1\text{:}T$. Since there are more voxels than TRs in most of the fMRI studies, $\mathbf{X}^{(i)}$ and the voxel correlation map $(\mathbf{X}^{(i)})^\top \mathbf{X}^{(j)}$ may not be full rank \cite{chen14,conroy09,lorbert12,xu12}. In addition, time synchronized stimulus ensures temporal alignment, i.e. the $m-th$ time point for all of the subjects represents the same simulation \cite{xu12,lorbert12}. Indeed, the main goal of HA methods is aligning the columns of $\mathbf{X}^{(i)}$ across subjects \cite{conroy09,xu12}, where the column representation of the functional activities for $i-th$ subject and $n-th$ voxel can be also defined as follows:
		\begin{equation}\label{eq:ColumnRep}
			\mathbf{x}^{(i)}_{.n}\in\mathbb{R}^T=\Big\{\mathbf{x}^{(i)}_{mn} | \mathbf{x}^{(i)}_{mn} \in \mathbf{X}^{(i)} \text{ and } m=1\text{:}T\Big\}
		\end{equation}
We firstly need a metric to qualify the functional alignment. Inter-Subject Correlation (ISC) is a classical metric for functional alignment, which can be defined for two different subjects as follows \cite{haxby11,xu12,chen14}:
		\begin{equation}\label{eq:ISCfunction}
			\begin{split}
				\text{ISC}(\mathbf{X}^{(i)},\mathbf{X}^{(j)})=(\sfrac{1}{V})\text{tr}((\mathbf{X}^{(i)})^\top \mathbf{X}^{(j)})=\\
				\frac{1}{V}\sum_{n=1}^{V}\big(\mathbf{x}^{(i)}_{.n} \big)^\top\mathbf{x}^{(j)}_{.n} = \frac{1}{V}\sum_{m=1}^{V}\sum_{n=1}^{V}\mathbf{x}^{(i)}_{mn}\mathbf{x}^{(j)}_{mn}
			\end{split}
		\end{equation}
where $\text{tr}()$ is the trace function. If the functional activities are column-wise standardized ($\mathbf{X}^{(i)} \sim \mathcal{N}(0,1)$), the ISC lies in $[-1, +1]$, where the large values represent better alignment \cite{conroy09,lorbert12,xu12,chen14}. Based on \eqref{eq:ISCfunction}, the hyperalignment can be formulated as follows:
		\begin{equation}
			\begin{split}\label{eq:ISCObjectiveFunction}
				\rho = \underset{i,j = 1:S}{\arg\max}\sum_{i<j}\text{ISC}(\mathbf{X}^{(i)}\mathbf{R}^{(i)},\mathbf{X}^{(j)}\mathbf{R}^{(j)})\\
				= \underset{i,j = 1:S}{\arg\max}\sum_{i<j}\sum_{m=1}^{V}\sum_{n=1}^{V}\mathbf{x}^{(i)}_{mn}\mathbf{r}^{(i)}_{nm}\mathbf{x}^{(j)}_{mn}\mathbf{r}^{(j)}_{nm}  
			\end{split}
		\end{equation}
where $\mathbf{R}^{(i)}=\Big\{\mathbf{r}^{(i)}_{mn} \Big\}\in\mathbb{R}^{V \times V}$ is the HA solution for $i-th$ subject. Constrains must be imposed in $\mathbf{R}^{(i)}$ to avoid overfitting \cite{xu12}. The general assumption in the basic hyperalignment is that the $\mathbf{R}^{(i)}, i=1\text{:}S$ are noisy `rotation' of a common template \cite{haxby11,guntupalli16,xu12}. This assumption leads us to define following problem:
\begin{equation}
\begin{split}\label{eq:HA}
\rho = \underset{i,j = 1:S}{\arg\min}\sum_{i<j} \| \mathbf{X}^{(i)}\mathbf{R}^{(i)} - \mathbf{X}^{(j)}\mathbf{R}^{(j)} \|^2_F\qquad \\
\text{subject to \quad} (\mathbf{R}^{(\ell)})^\top \mathbf{A}^{(\ell)}\mathbf{R}^{(\ell)}=\mathbb{I},\qquad \ell=1\text{:}S
\end{split}
\end{equation}
where $\mathbb{I}$ denotes the identity matrix and the matrices $\mathbf{A}^{(\ell)} \in \mathbb{R}^{V \times V}, \ell = 1\text{:}S$ are symmetric and positive definite. Generally, if $\mathbf{A}^{(\ell)} = \mathbb{I}$, then we have hyperalignment or a multi-set orthogonal Procrustes problem, which is commonly used in share analysis. In addition, if $\mathbf{A}^{(\ell)} = (\mathbf{X}^{(\ell)})^\top \mathbf{X}^{(\ell)}$, then \eqref{eq:HA} denotes a form of multi-set Canonical Correlation Analysis (CCA) \cite{lorbert12,xu12,lorbert12,chen14}. 
\begin{lemma}
\emph{The equation \eqref{eq:HA} is equivalent to:}
\begin{equation}
\begin{split}\label{eq:CCA}
\rho = \underset{}{\arg\min}\sum_{i=1}^{S} \| \mathbf{X}^{(i)}\mathbf{R}^{(i)} - \mathbf{G} \|^2_F \qquad \\
\text{subject to \qquad} (\mathbf{R}^{(\ell)})^\top \mathbf{A}^{(\ell)}\mathbf{R}^{(\ell)}=\mathbb{I},\qquad \ell=1\text{:}S
\end{split}
\end{equation}
\emph{where $\mathbf{G} \in \mathbb{R}^{T\times V}$ is the HA template:}
\begin{equation}
\begin{split}\label{eq:G}
\mathbf{G} = \frac{1}{S} \sum_{j=1}^{S} \mathbf{X}^{(j)}\mathbf{R}^{(j)}
\end{split}
\end{equation}	
\emph{Proof.} Please refer to \cite{gower04,lorbert12} for the proof.
\end{lemma}
Indeed, the HA template ($\mathbf{G}$) can be used for functional alignment in the test-set before MVP analysis. Most of previous studies have used CCA for finding this template \cite{xu12,haxby11,chen14}. 
\begin{lemma}\label{lm:CCA}
\emph{Canonical Correlation Analysis (CCA) finds an optimum solution for solving \eqref{eq:HA} by exploiting the objective function $ \underset{i,j = 1\text{:}S}{\max}\Big((\mathbf{R}^{(i)})^\top \mathbf{C}^{(i,j)} \mathbf{R}^{(j)}\Big)$, and then $\mathbf{G}$ also can be calculated based on \eqref{eq:G}. Briefly, the CCA solution can be formulated as follows: 
\begin{equation}\label{eq:CCASol}
\rho = \underset{i,j=1\text{:}S}{\arg\max}\bigg(\frac{(\mathbf{R}^{(i)})^\top \mathbf{C}^{(i,j)} \mathbf{R}^{(j)}}{\sqrt{((\mathbf{R}^{(i)})^\top \mathbf{C}^{(i)}\mathbf{R}^{(i)})((\mathbf{R}^{(j)})^\top \mathbf{C}^{(j)}\mathbf{R}^{(j)})}}\bigg)
\end{equation}
where $\mathbf{C}^{(i)}\in\mathbb{R}^{V \times V}=\mathbb{E}\Big[(\mathbf{X}^{(i)})^\top \mathbf{X}^{(i)}\Big]=(\mathbf{X}^{(i)})^\top \mathbf{X}^{(i)}$, $\mathbf{C}^{(j)}\in\mathbb{R}^{V \times V}=\mathbb{E}\Big[(\mathbf{X}^{(j)})^\top \mathbf{X}^{(j)}\Big]=(\mathbf{X}^{(j)})^\top \mathbf{X}^{(j)}$, and $\mathbf{C}^{(i,j)}\in\mathbb{R}^{V \times V}=\mathbb{E}\Big[(\mathbf{X}^{(i)})^\top \mathbf{X}^{(j)}\Big]=(\mathbf{X}^{(i)})^\top \mathbf{X}^{(j)}$. The solution of CCA can be obtained by computing a generalized eigenvalue decomposition problem} \cite{hardoon04,peng10}.\\
\emph{Proof.} Equation \eqref{eq:HA} can be written as follows:
\begin{equation}
\begin{split}
\| \mathbf{X}^{(i)}\mathbf{R}^{(i)} - \mathbf{X}^{(j)}\mathbf{R}^{(j)} \|^2_F\ = -2(\mathbf{R}^{(i)})^\top (\mathbf{X}^{(i)})^\top \mathbf{X}^{(j)} \mathbf{R}^{(j)}\\+(\mathbf{R}^{(i)})^\top (\mathbf{X}^{(i)})^\top \mathbf{X}^{(i)} \mathbf{R}^{(i)} + (\mathbf{R}^{(j)})^\top (\mathbf{X}^{(j)})^\top \mathbf{X}^{(j)} \mathbf{R}^{(j)}\equiv\\\frac{(\mathbf{R}^{(i)})^\top(\mathbf{X}^{(i)})^\top \mathbf{X}^{(j)} \mathbf{R}^{(j)}}{\sqrt{((\mathbf{R}^{(i)})^\top (\mathbf{X}^{(i)})^\top \mathbf{X}^{(i)}\mathbf{R}^{(i)})((\mathbf{R}^{(j)})^\top (\mathbf{X}^{(j)})^\top \mathbf{X}^{(j)}\mathbf{R}^{(j)})}}
\end{split}
\end{equation}
\end{lemma}
\begin{remark}
\emph{The HA solution generated by unsupervised CCA may not be optimum for MVP analysis. We just explain two issues in the unsupervised solutions. Consider fMRI time series included visual stimuli, where two subjects watch two photos of cats as well as two photos of human faces. In this example, the sequence of stimuli is demonstrated by $[\mathbf{cat1}, \mathbf{face1}, \mathbf{cat2}, \mathbf{face2}]$ for each subject after preprocessing steps. The unsupervised solution finds two mappings to maximize the correlation in the voxel-level, where the voxels for each subject are only compared with the voxels for other subjects with the same locations. As a result, the sequence of the mentioned comparison in the stimulus-level is shown by:
\begin{equation*}
\begin{split}
(\mathbf{S1\text{:}cat1\uparrow S2\text{:}cat1})\text{ ; }(\mathbf{S1\text{:}face1\uparrow S2\text{:}face1})\text{;}\\
(\mathbf{S1\text{:}cat2\uparrow S2\text{:}cat2})\text{ ; }(\mathbf{S1\text{:}face2\uparrow S2\text{:}face2})
\end{split}
\end{equation*}
where $\uparrow$ denotes the operator for maximizing correlation and the $\mathbf{S1}$ and $\mathbf{S2}$ are the indices of subjects 1 and 2, respectively. Now, we can explain two issues. Indeed, the CCA solution here just maximized the correlation for the stimuli in the same locations, while they must also maximize the correlation between all stimuli in the same category and minimize the correlation between different categories of stimuli. Our approach for solving mentioned issues can be illustrated by:
\begin{equation*}
\begin{split}
(\mathbf{S1\text{:}cat1,2\uparrow S2\text{:}cat1,2})\text{;  }(\mathbf{S1\text{:}face1,2\uparrow S2:face1,2})\text{;}\\
(\mathbf{S1\text{:}cat1,2\downarrow S2\text{:}face1,2})\text{;  }(\mathbf{S1\text{:}face1,2\downarrow S2\text{:}cat1,2})
\end{split}
\end{equation*}
where $\downarrow$ denotes the operator for minimizing correlation.}
\end{remark}		
This paper proposes Local Discriminant Hyperalignment (LDHA), which combines the idea of locality into CCA \cite{peng10} in order to provide a better HA solution in the MVP analysis. Since unaligned (before applying the HA method) functional activities in different subjects cannot be directly compared with each other, the neighborhoods matrix $\mathbf{\alpha}=\Big\{ \mathbf{\alpha}_{mn}\Big\}\in\mathbb{R}^{T\times T}$ is defined by using class labels ($\mathbf{Y}$) in the train-set as follows:
\begin{equation}\label{eq:Alpha}
\mathbf{\alpha}_{nm}=\mathbf{\alpha}_{mn}=\begin{cases}
0 & \quad \mathbf{y}_m \neq \mathbf{y}_n\\
1 & \quad \mathbf{y}_m = \mathbf{y}_n
\end{cases},\quad m,n=1\text{:}T\text{, } m<n
\end{equation} 
where the number of within-class neighborhoods is the permutation of all stimuli in each category, and the number of between-classes neighborhoods denotes by the permutation of all stimuli in distinct categories. The covariance matrices for within-class $\mathbf{W}^{(i,j)} = \Big\{\mathbf{w}^{(i,j)}_{mn}\Big\} \in \mathbb{R}^{V\times V}$ and between-classes $\mathbf{B}^{(i,j)} = \Big\{\mathbf{b}^{(i,j)}_{mn}\Big\} \in \mathbb{R}^{V\times V}$ are defined as follows: 
\begin{equation}\label{eq:withinclass}
\mathbf{w}^{(i,j)}_{mn} = \sum_{\ell=1}^{T}\sum_{k=1}^{T} \mathbf{\alpha}_{\ell k}\mathbf{x}_{\ell m}^{(i)}\mathbf{x}_{kn}^{(j)} + \mathbf{\alpha}_{\ell k}\mathbf{x}_{\ell n}^{(i)}\mathbf{x}_{k m}^{(j)}
\end{equation}
\begin{equation}\label{eq:betweenclasses}
\mathbf{b}^{(i,j)}_{mn} = \sum_{\ell=1}^{T}\sum_{k=1}^{T} (1 - \mathbf{\alpha}_{\ell k})\mathbf{x}_{\ell m}^{(i)}\mathbf{x}_{kn}^{(j)} + (1 - \mathbf{\alpha}_{\ell k})\mathbf{x}_{\ell n}^{(i)}\mathbf{x}_{k m}^{(j)}
\end{equation}
where $m,n = 1\text{:}V$. The Local Discriminant Hyperalignment (LDHA) objective function is defined by $ \underset{i,j = 1\text{:}S}{\max}\Big((\mathbf{R}^{(i)})^\top \widetilde{\mathbf{C}}^{(i,j)} \mathbf{R}^{(j)}\Big)$ where $\widetilde{\mathbf{C}}^{(i,j)} = \mathbf{W}^{(i,j)}- (\eta / T^2) \mathbf{B}^{(i,j)}$. Here, $\eta$ is the number of non-zero cells in the matrix $\alpha$, and $T$ is the number of time points in unites of TRs. In addition, the solution of the LDHA can be reformulated as follows:
\begin{equation}
\begin{split}\label{eq:LDHA_ObF}
\rho = \underset{i,j=1\text{:}S, i<j}{\arg\max}\frac{(\mathbf{R}^{(i)})^\top \widetilde{\mathbf{C}}^{(i,j)} \mathbf{R}^{(j)}}{\sqrt{((\mathbf{R}^{(i)})^\top \mathbf{C}^{(i)}\mathbf{R}^{(i)})((\mathbf{R}^{(j)})^\top \mathbf{C}^{(j)}\mathbf{R}^{(j)})}}\\
\text{subject to \quad} (\mathbf{R}^{(\ell)})^\top \mathbf{C}^{(\ell)}\mathbf{R}^{(\ell)} = \mathbb{I}\text{,} \qquad\quad \ell = 1\text{:}S
\end{split}
\end{equation}	
The main difference between LHDA and HA is the supervised covariance matrix ($\widetilde{\mathbf{C}}^{(i,j)}$). Indeed, LDHA can be considered equivalent of the classical CCA (Lemma \ref{lm:CCA}), where the correlations of non-homogeneous stimuli ($\mathbf{B}^{(i,j)}$) are participated to the CCA problem with a negative sign and all of the homogeneous stimuli in each category will be compared ($\mathbf{W}^{(i,j)}$) with each other. In addition, LDHA is related to the LDCCA method \cite{peng10}, where we have a mechanism in the LDCCA to manually select the relevant neighborhoods for each category of stimuli based on class-labels and also the balance factor is dynamically assigned based on the data structure.
\begin{lemma}
\emph{Same as the classical CCA, LDHA can be solved as a generalized eigenvalue decomposition problem.\\Proof.}
\begin{equation}
\begin{split}\label{eq:LDHA2}
\widetilde{\mathbf{C}}^{(i,j)}\Big(\mathbf{C}^{(j)}\Big)^{-1}\widetilde{\mathbf{C}}^{(j,i)}\mathbf{R}^{(i)}=\Big(\mathbf{\Lambda}^{(i,j)}\Big)^2 \mathbf{C}^{(i)}\mathbf{R}^{(i)}\\
\widetilde{\mathbf{C}}^{(j,i)}\Big(\mathbf{C}^{(i)}\Big)^{-1}\widetilde{\mathbf{C}}^{(i,j)}\mathbf{R}^{(j)}=\Big(\mathbf{\Lambda}^{(i,j)}\Big)^2 \mathbf{C}^{(j)}\mathbf{R}^{(j)}
\end{split}
\end{equation}
\emph{This paper uses the Singular Value Decomposition (SVD) to solve LDHA problem, where $\mathbf{H}^{(i,j)} = \Big(\mathbf{C}^{(i)}\Big)^\mathbf{\sfrac{-1}{2}}\widetilde{\mathbf{C}}^{(i,j)}\Big(\mathbf{C}^{(j)}\Big)^{\mathbf{\sfrac{-1}{2}}}$, $\mathbf{P}^{(i,j)} = \Big(\mathbf{C}^{(i)}\Big)^\mathbf{\sfrac{1}{2}}\mathbf{R}^{(i)}$, and $\mathbf{Q}^{(i,j)} = \Big(\mathbf{C}^{(j)}\Big)^\mathbf{\sfrac{1}{2}}\mathbf{R}^{(j)}$. By considering SVD parameters, the \eqref{eq:LDHA2} is equivalent to:}
\begin{equation}
\begin{cases}
\mathbf{H}^{(i,j)}\Big(\mathbf{H}^{(i,j)}\Big)^\top \mathbf{P}^{(i,j)} = \Big(\mathbf{\Lambda}^{(i,j)}\Big)^2 \mathbf{P}^{(i,j)}\\
\Big(\mathbf{H}^{(i,j)}\Big)^\top \mathbf{H}^{(i,j)}\mathbf{Q}^{(i,j)}= \Big(\mathbf{\Lambda}^{(i,j)}\Big)^2 \mathbf{Q}^{(i,j)}
\end{cases}
\end{equation}   
\emph{where $\mathbf{H}^{(i,j)}=\mathbf{P}^{(i,j)}\mathbf{\Lambda}^{(i,j)}\Big(\mathbf{Q}^{(i,j)}\Big)^\top$. The Hyperalignment solution also can be defined as follows:
\begin{equation}
\begin{cases}
\mathbf{R}^{(i)}=\Big(\mathbf{C}^{(i)}\Big)^\mathbf{\sfrac{-1}{2}}\mathbf{P}^{(i,j)}\\
\mathbf{R}^{(j)}=\Big(\mathbf{C}^{(j)}\Big)^\mathbf{\sfrac{-1}{2}}\mathbf{Q}^{(i,j)}
\end{cases}
\end{equation}	
}
\end{lemma}
Algorithm \ref{alg:LDHA} illustrates the LDHA procedure for solving the HA problem between two different subjects. As mentioned before, the LDHA is used for MVP analysis. Algorithm \ref{alg:MVP} demonstrates a general template for MVP analysis based on LDHA method. As this algorithm depicted, the procedure of generating the HA template ($\mathbf{G}$) in the train stage is changed, while the template is used in the test stage such as the unsupervised HA methods. Therefore, we \textbf{do not} need the class labels in the test stage. Indeed, the proposed method in comparison with the unsupervised solutions just generates more optimum HA template for aligning functional neural activities, where this template can maximize the correlation between all stimuli in the same category and minimize the correlation between different categories of stimuli.  
\begin{algorithm}[!t]
			\caption{Local Discriminate Hyperalignment (LDHA)}
			\label{alg:LDHA}
			\begin{algorithmic}
				\STATE {\bfseries Input:} Data points $\mathbf{X}^{(i)}$ and $\mathbf{X}^{(j)}$, class labels $\mathbf{Y}$:\\
				\STATE {\bfseries Output:} Hyperalignment parameters $\mathbf{R}^{(i)}$ and $\mathbf{R}^{(j)}$:\\
				\STATE {\bfseries Method:}\\
				\quad1. Generate $\alpha$ by (\ref{eq:Alpha}).\\
				\quad2. Calculate $\mathbf{W}^{(i,j)}$, $\mathbf{B}^{(i,j)}$ by using (\ref{eq:withinclass}) and (\ref{eq:betweenclasses}).\\
				\quad3. Calculate $\widetilde{\mathbf{C}}^{(i,j)}$.\\
				\quad4. Compute $\mathbf{H}^{(i,j)} = \Big(\mathbf{C}^{(i)}\Big)^\mathbf{\sfrac{-1}{2}}\widetilde{\mathbf{C}}^{(i,j)}\Big(\mathbf{C}^{(j)}\Big)^\mathbf{\sfrac{-1}{2}}$.\\
				\quad5. Perform SVD: $\mathbf{H}^{(i,j)}=\mathbf{P}^{(i,j)}\mathbf{\Lambda}^{(i,j)}\Big(\mathbf{Q}^{(i,j)}\Big)^\top$.\\
				\quad6. Return $\mathbf{R}^{(i)}=\Big(\mathbf{C}^{(i)}\Big)^\mathbf{\sfrac{-1}{2}}\mathbf{P}^{(i,j)}$\\
				\quad\quad\quad \textbf{and   } $\mathbf{R}^{(j)}=\Big(\mathbf{C}^{(j)}\Big)^{\sfrac{-1}{2}}\mathbf{Q}^{(i,j)}$.
			\end{algorithmic}
\end{algorithm}
\begin{algorithm}[!t]
			\caption{A general template for MVP analysis by using Local Discriminate Hyperalignment (LDHA)}
			\label{alg:MVP}
			\begin{algorithmic}
				\STATE {\bfseries Input:} Train Set $\mathbf{X}^{(i)}, i=1\text{:}S$, Test Set $\widehat{\mathbf{X}}^{(j)}, j=1\text{:}\hat{S}$:\\
				\STATE {\bfseries Output:} Classification Performance ($ACC$, $AUC$):\\
				\STATE {\bfseries Method:}\\
				\quad01. Initiate $\mathbf{R}^{(i)}\text{,  } i=1\text{:}S$.\\
				\quad02. \textbf{Do}\\
				\quad03. {\bfseries Foreach} subject $\mathbf{X}^{(i)}, i=1\text{:}S$:\\
				\quad04. Update $\mathbf{R}^{(i)}$ by Alg. \ref{alg:LDHA} and $\mathbf{X}^{(\ell)}, \ell=i\text{+}1\text{:}S$.\\
				\quad05. {\bfseries End Foreach}\\
				\quad06. \textbf{Until} $\mathbf{X}^{(i)}\mathbf{R}^{(i)}\text{,  } i=1\text{:}S$ do not change in this step.\\
				\quad07. Train a classifier by $\mathbf{X}^{(i)}\mathbf{R}^{(i)}, i=1\text{:}S$\\
				\quad08. Initiate $\widehat{\mathbf{R}}^{(j)}\text{,  } j=1\text{:}\hat{S}$.\\
				\quad09. Generate $\mathbf{G}$ based on \eqref{eq:G} by using $\mathbf{R}^{(i)}, i=1\text{:}S$\\				
				\quad10. {\bfseries Foreach} subject $\widehat{\mathbf{X}}^{(j)}, j=1\text{:}\hat{S}$:\\
				\quad11. Compute $\widehat{\mathbf{R}}^{(j)}$ by \emph{classical} HA (Eq. \ref{eq:CCA},\ref{eq:CCASol}) and $\textbf{G}$.\\
				\quad12. {\bfseries End Foreach}\\
				\quad13. Evaluate the classifier by using $\widehat{\mathbf{X}}^{(j)}\widehat{\mathbf{R}}^{(j)}, j=1\text{:}\hat{S}$.
			\end{algorithmic}
\end{algorithm}
\begin{table*}
			\caption{Accuracy of Classification Methods}
			\label{tbl:BinaryAccuracy}
			\begin{center}
				\begin{small}
					\begin{tabular}{lcccccc}
						\hline
						Data Sets & $\nu$-SVM & HA & KHA & SCCA & SVD-HA & LDHA \\
						\hline
						DS005 ($2$ classes)& 71.65$\pm$0.97 & 81.27$\pm$0.59 & 83.06$\pm$0.36 & 85.29$\pm$0.49 & 90.82$\pm$1.23 & \textbf{94.32$\pm$0.16}  \\
						DS105 ($8$ classes)& 22.89$\pm$1.02 & 30.03$\pm$0.87 & 32.62$\pm$0.52 & 37.14$\pm$0.91 & 40.21$\pm$0.83 & \textbf{54.04$\pm$0.09}  \\
						DS107 ($4$ classes)& 38.84$\pm$0.82 & 43.01$\pm$0.56 & 46.82$\pm$0.37 & 52.69$\pm$0.69 & 59.54$\pm$0.99 & \textbf{74.73$\pm$0.19}\\
						DS117 ($2$ classes)& 73.32$\pm$1.67 & 77.93$\pm$0.29 & 84.22$\pm$0.44 & 83.32$\pm$0.41 & \textbf{95.62$\pm$0.83} & 95.07$\pm$0.27\\		
						\hline
					\end{tabular}
				\end{small}
			\end{center}
			\vskip -0.25in
		\end{table*}
		\begin{table*}
			\caption{Area Under the ROC Curve (AUC) of Classification Methods  }
			\label{tbl:BinaryAUC}
			\begin{center}
				\begin{small}
					\begin{tabular}{lcccccc}
						\hline
						Data Sets & $\nu$-SVM & HA & KHA & SCCA & SVD-HA & LDHA \\
						\hline
						DS005 ($2$ classes) & 68.37$\pm$1.01 & 70.32$\pm$0.92 & 82.22$\pm$0.42 & 80.91$\pm$0.21 & 88.54$\pm$0.71 & \textbf{93.25$\pm$0.92}  \\
						DS105 ($8$ classes)& 21.76$\pm$0.91 & 28.91$\pm$1.03 & 30.35$\pm$0.39 & 36.23$\pm$0.57 & 37.61$\pm$0.62 & \textbf{53.86$\pm$0.17}  \\
						DS107 ($4$ classes)& 36.84$\pm$1.45 & 40.21$\pm$0.33 & 43.63$\pm$0.61 & 50.41$\pm$0.92 & 57.54$\pm$0.31 & \textbf{72.03$\pm$0.37}\\
						DS117 ($2$ classes)& 70.17$\pm$0.59 & 76.14$\pm$0.49 & 81.54$\pm$0.92 & 80.92$\pm$0.28 & 92.14$\pm$0.42 & \textbf{94.23$\pm$0.94}\\
						\hline
					\end{tabular}
				\end{small}
			\end{center}
			\vskip -0.2in
		\end{table*}		
\section{Experiments}
The empirical studies are presented in this section. Same as previous studies \cite{chen14,xu12,lorbert12,haxby11}, this paper generates the classification model by using the $\nu$-SVM algorithms \cite{smola04}, i.e. the binary $\nu$-SVM for datasets with just two categories of stimuli, and multi-label $\nu$-SVM \cite{smola04,lorbert12} as multi-class approach. All employed datasets in this paper are separately preprocessed by SPM 12 (6685) (www.fil.ion.ucl.ac.uk/spm/), i.e. slice timing, anatomical alignment, normalization, smoothing. Regions of Interests (ROIs) are also defined by using the main references of each dataset. The features (voxels in the ROIs) are partitioned to train set and test set by using Leave-One-Out (LOO) cross-validation across subjects (leave-one-subject-out). The HA methods are applied for functional aligning the neural activities and generating the general template $(\mathbf{G})$. Finally, the classification model is generated for evaluating the performance of different methods. Performance of LDHA is compared with the $\nu$-SVM algorithm as the baseline (it just uses anatomical alignment without the hyperalignment mapping), the standard hyperalignment (HA) \cite{haxby11,guntupalli16}, Kernel-based hyperalignment (KHA) \cite{lorbert12}, Regularized hyperalignment (SCCA) \cite{xu12}, and Joint SVD hyperalignment (SVD-HA) \cite{chen14}. Further, KHA algorithm is employed by the Gaussian kernel, which generated the best results in the original paper \cite{lorbert12}. In addition, regularized parameters ($\alpha,\beta$) in SCCA are considered optimum based on \cite{xu12}. All algorithms are implemented in the MATLAB R2016b (9.1) on a PC with certain specifications\footnote{DEL , CPU = Intel Xeon E5-2630 v3 (8$\times$2.4 GHz), RAM = 64GB, OS = Elementary OS 0.4 Loki} by authors in order to generate experimental results. 
\subsection{Simple Tasks Analysis}
This paper utilizes 4 datasets, shared by openfmri.org, for running empirical studies of this section. These datasets contain simple tasks such as watching a gray-scale photo or tapping a key, etc. As the first dataset, `Visual Object Recognition' (DS105) includes $6$ subjects and $71$ sessions. It also contains $8$ classes (categories) of visual stimuli, i.e. gray-scale images of faces, houses, cats, bottles, scissors, shoes, chairs, and scrambles (nonsense patterns). Please see \cite{haxby11,haxby14} for more information. As the second dataset, `Multi-subject, multi-modal human neuroimaging dataset' (DS117) includes MEG and fMRI images for $19$ subjects and $171$ sessions. This paper just uses the fMRI images of this dataset. It also contains $2$ classes of visual stimuli, i.e. human faces, and scrambles. Please see \cite{wakeman15} for more information. The responses of voxels in the ventral temporal cortex (VT) are analyzed for these two datasets. As the third dataset, `Word and Object Processing' (DS107) includes $49$ subjects and $98$ sessions. It contains $4$ classes of visual stimuli, i.e. words, objects, scrambles, consonants. ROIs and technical information are defined based on \cite{duncan09}. As the last dataset, `Mixed-gambles task'  (DS005) includes $16$ subjects and $48$ sessions. It also contains $2$ classes of risk tasks in the human brain, where the chance of selection is $50/50$. Further, the ROIs for functional alignment are selected based on the original paper \cite{tom07}. 

Table \ref{tbl:BinaryAccuracy} and \ref{tbl:BinaryAUC} respectively demonstrate the classification Accuracy and Area Under the ROC Curve (AUC) in percentage (\%) for the predictors. These tables report the performance of predictors based on the categories of the stimuli. As these tables demonstrate, the proposed algorithm has generated better performance in comparison with other methods because it provided a better functional alignment of neural activities by exploiting the locality properties of the fMRI datasets. This issue is more significant when the number of classes is more than 2 such as datasets DS105 and DS107.
		\begin{figure}[!t]
			\begin{center}
				\begin{minipage}{0.48\linewidth}
					\includegraphics[width=0.98\textwidth,height=0.6\linewidth]{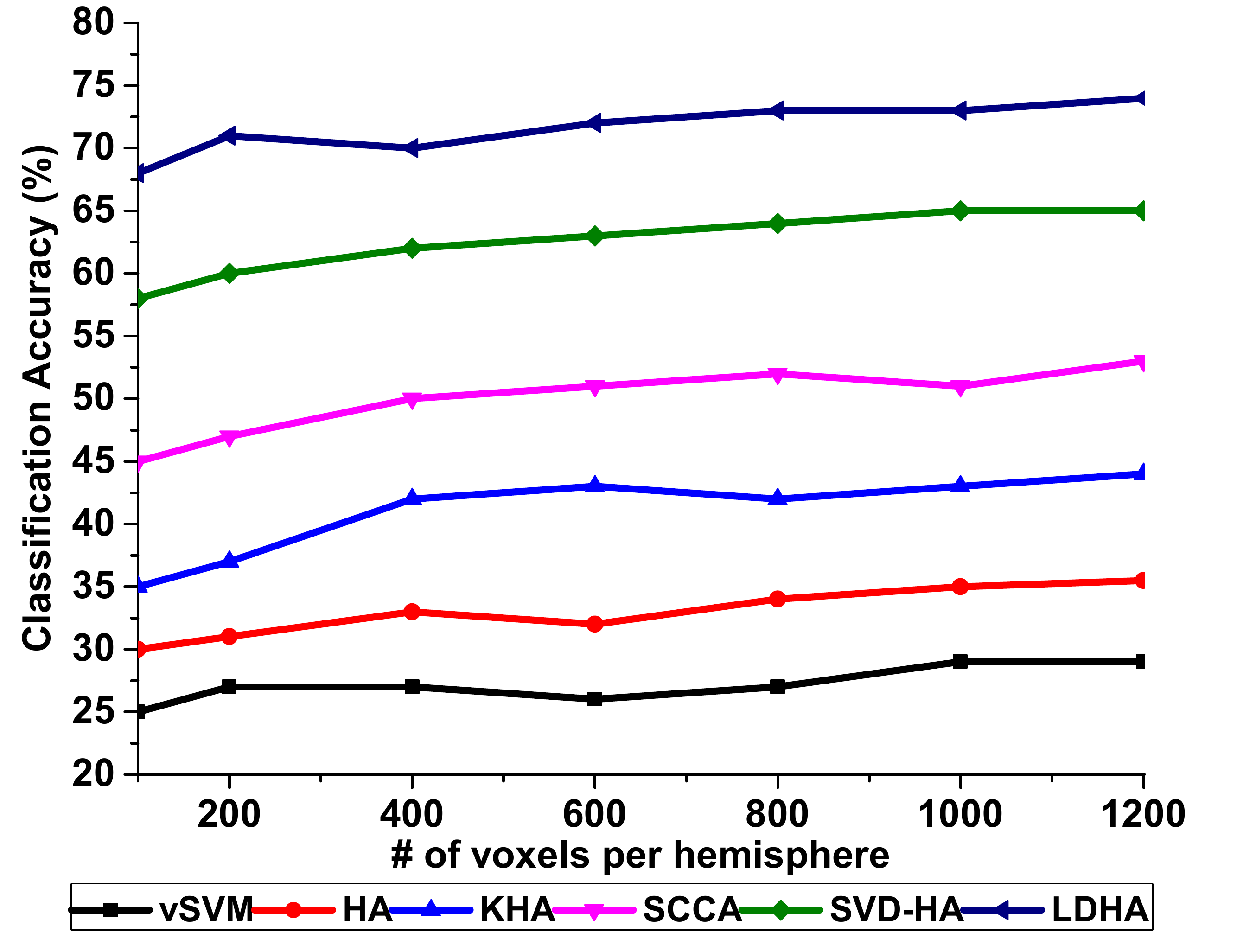}\\
					\centering (a) Forrest Gump\\(TRs = 100)
				\end{minipage}
				\begin{minipage}{0.48\linewidth}
					\includegraphics[width=0.98\textwidth,height=0.6\linewidth]{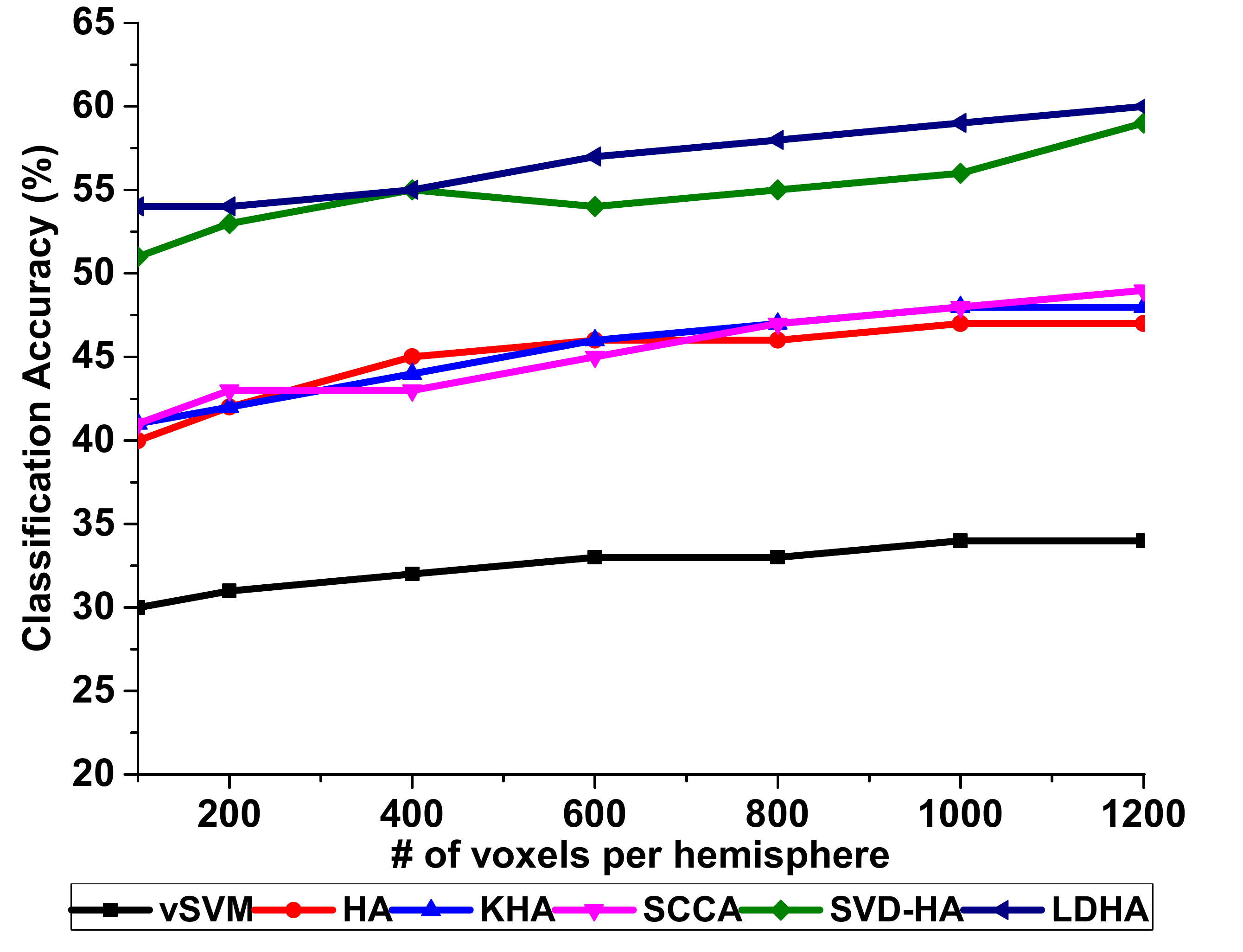}\\
					\centering (b) Raiders of the Lost Ark\\(TRs = 100)
				\end{minipage}	
				\begin{minipage}{0.48\linewidth}
					\includegraphics[width=0.98\textwidth,height=0.6\linewidth]{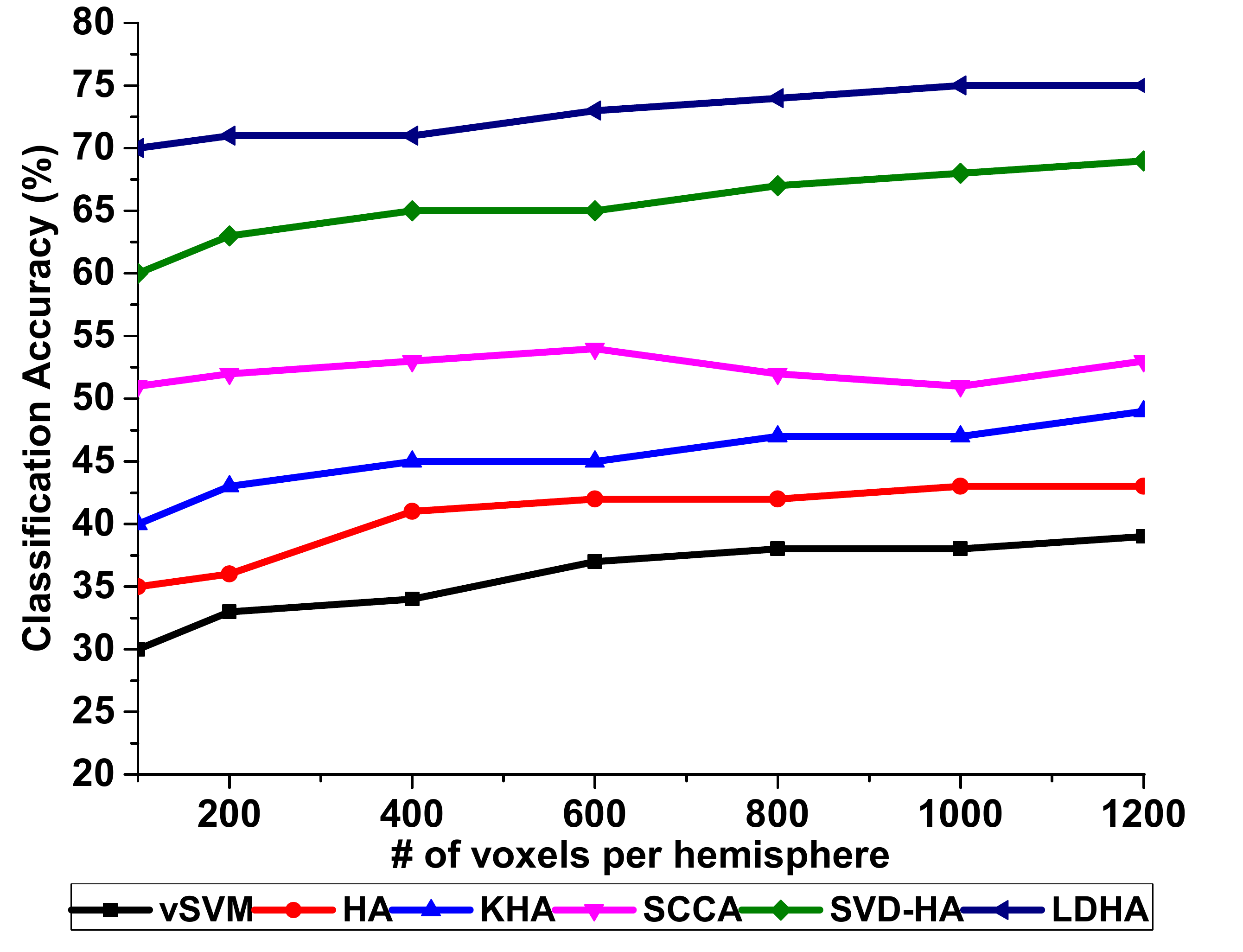}\\
					\centering (c) Forrest Gump\\(TRs = 200)
				\end{minipage}
				\begin{minipage}{0.48\linewidth}
					\includegraphics[width=0.98\textwidth,height=0.6\linewidth]{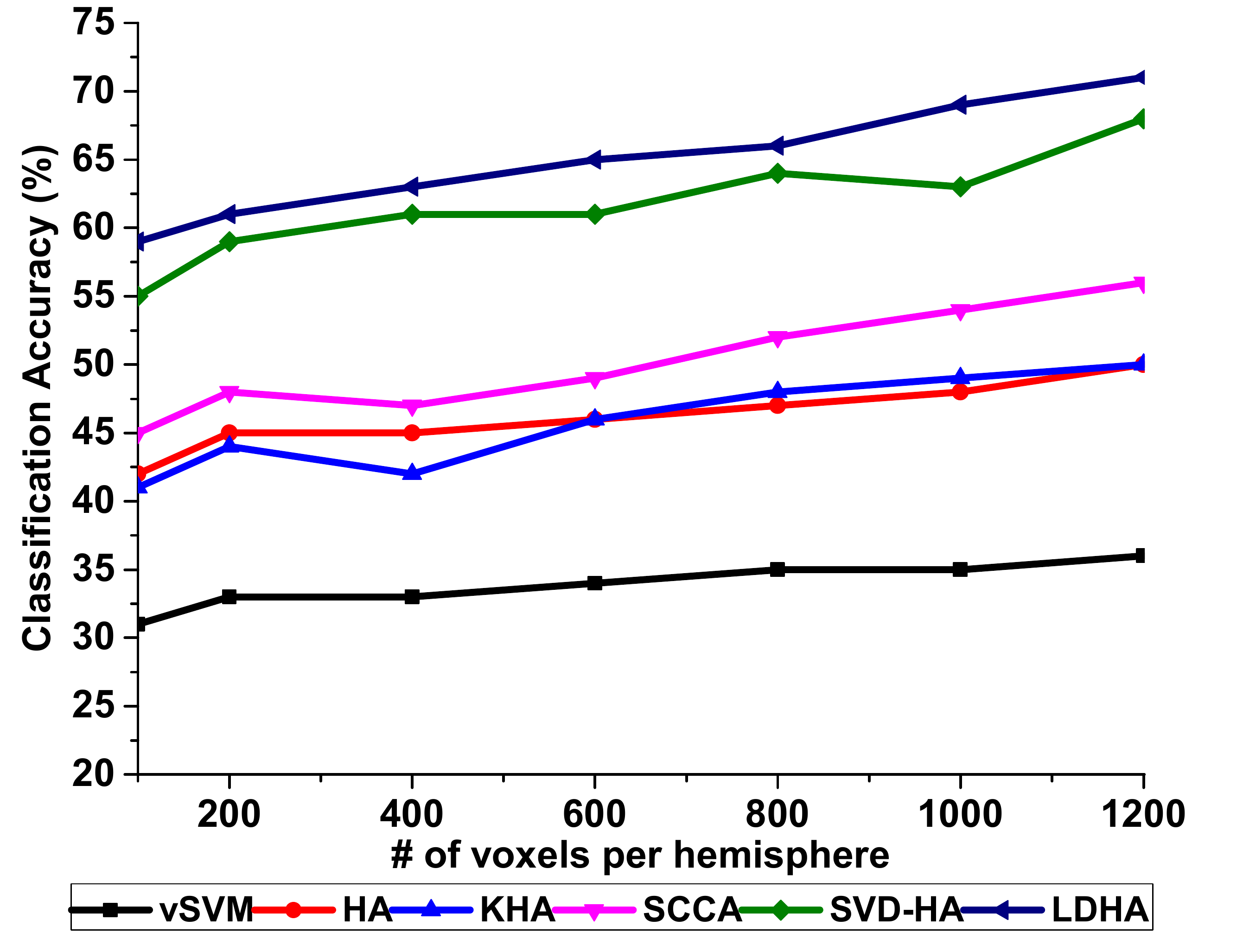}\\
					\centering (d) Raiders of the Lost Ark\\(TRs = 200)
				\end{minipage}			
				\begin{minipage}{0.48\linewidth}
					\includegraphics[width=0.98\textwidth,height=0.6\linewidth]{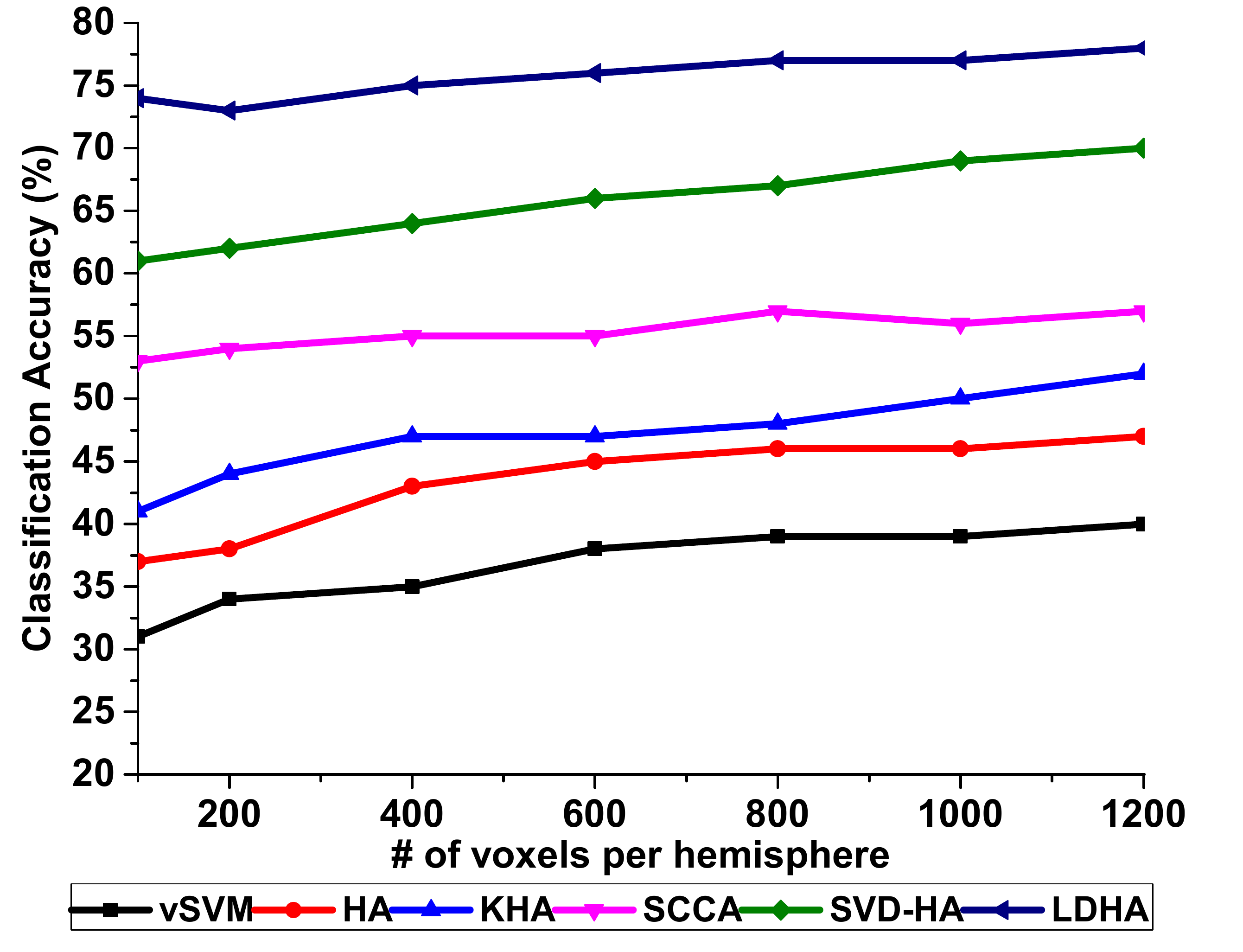}\\
					\centering (e) Forrest Gump\\(TRs = 400)
				\end{minipage}
				\begin{minipage}{0.48\linewidth}
					\includegraphics[width=0.98\textwidth,height=0.6\linewidth]{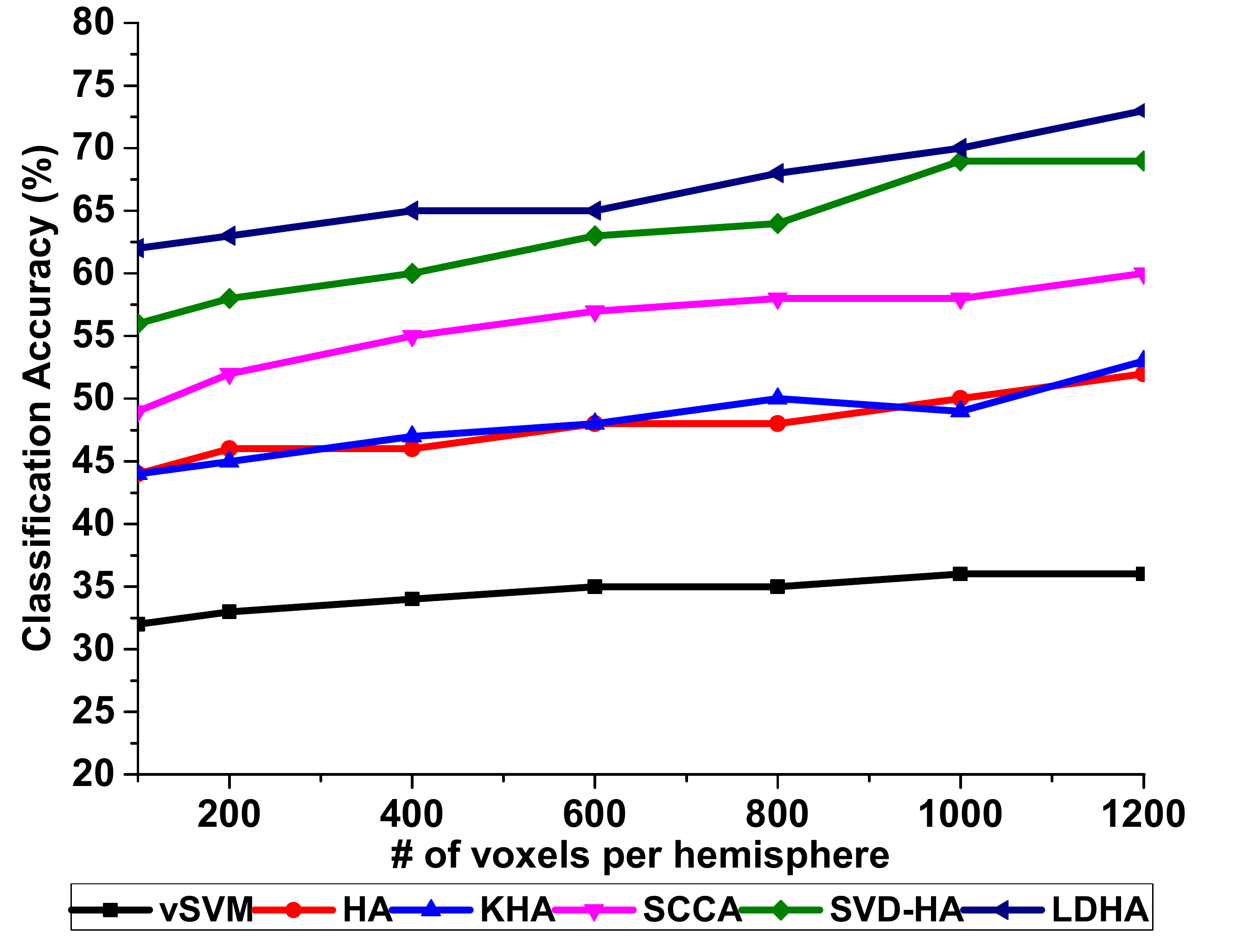}\\
					\centering (f) Raiders of the Lost Ark\\(TRs = 400)
				\end{minipage}				
				\begin{minipage}{0.48\linewidth}
					\includegraphics[width=0.98\textwidth,height=0.6\linewidth]{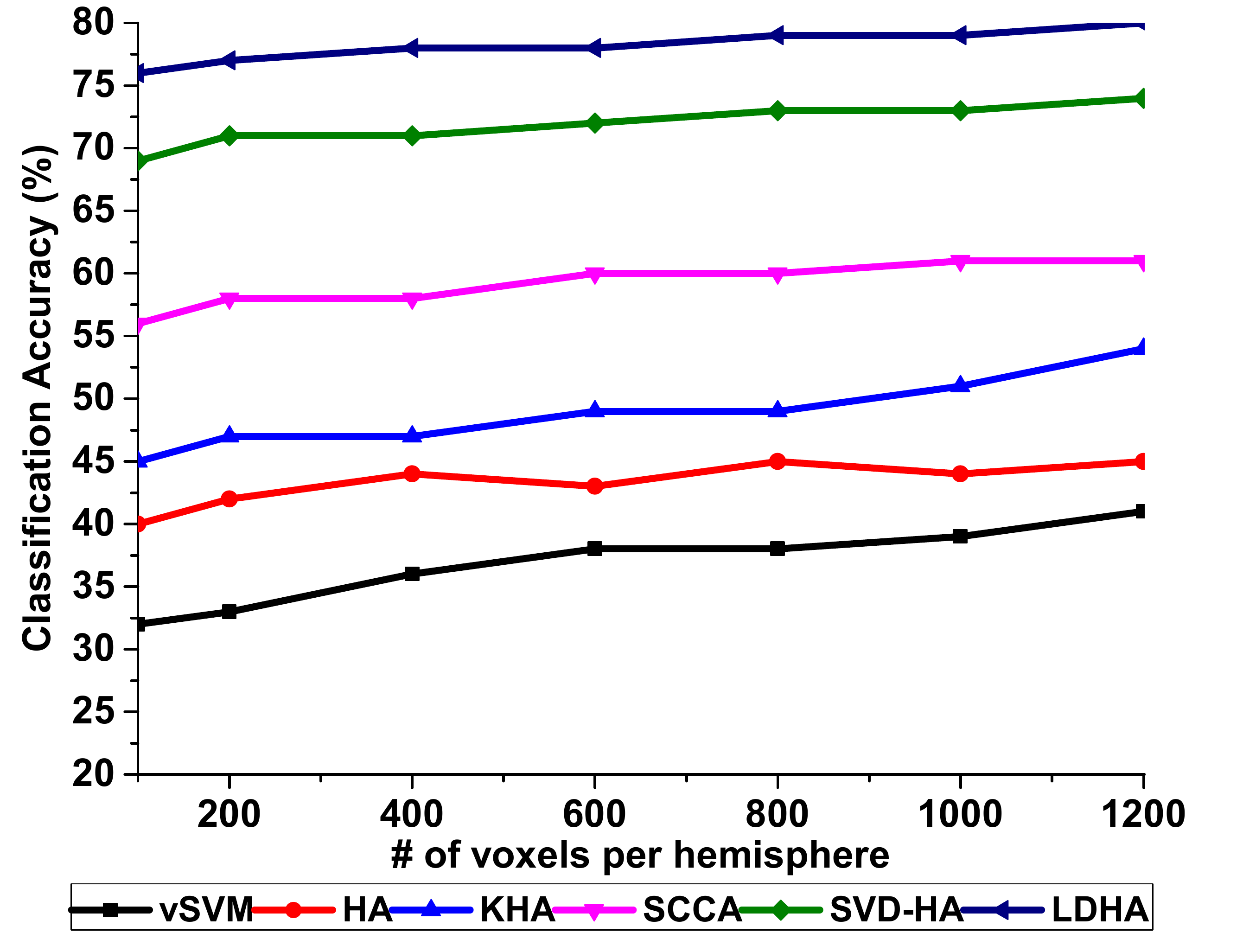}\\
					\centering (g) Forrest Gump\\(TRs = 2000)
				\end{minipage}
				\begin{minipage}{0.48\linewidth}
					\includegraphics[width=0.98\textwidth,height=0.6\linewidth]{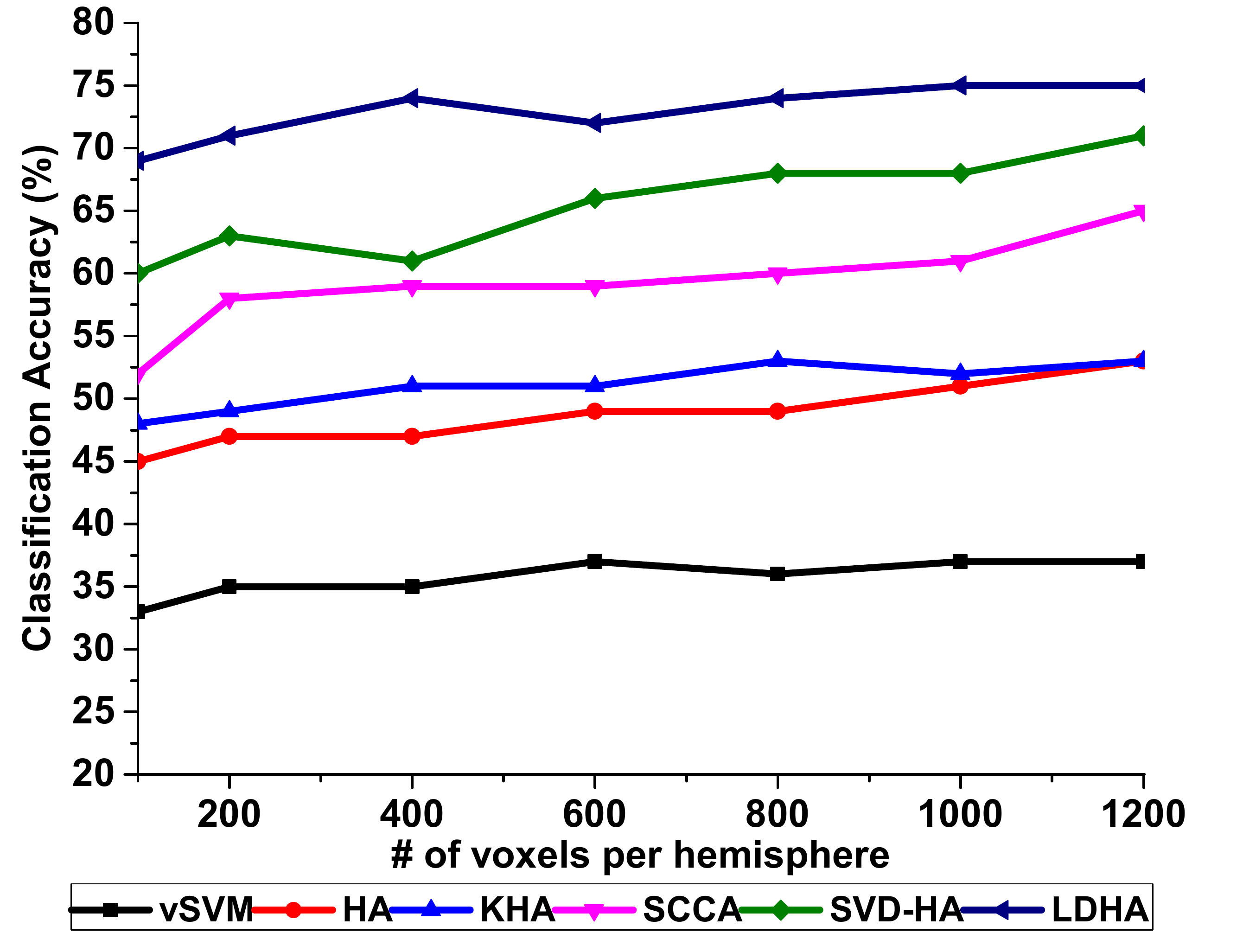}\\
					\centering (h) Raiders of the Lost Ark\\(TRs = 2000)
				\end{minipage}			
				\caption{Comparison of 6 different HA algorithms by using 100 TRs, 400 TRs, 800 TRs, and 2000 TRs and 100 to 1200 ranked voxels.}
				\label{fig:Movies}
			\end{center}
			\vskip -0.3in
		\end{figure}
\subsection{Complex Tasks Analysis}
This section employs two fMRI datasets, which are related to watching movies. As the first dataset, `A high-resolution 7-Tesla fMRI dataset from complex natural stimulation with an audio movie' (DS113) includes the fMRI data of 20 subjects, who watched `Forrest Gump (1994)' movie during the experiment. This dataset provided by www.openfmri.org. Please see \cite{hanke14} for more information. In the second dataset, subjects watched `Raiders of the Lost Ark (1981)', where whole brain volumes are 48 and the number of subjects are 10. Please see \cite{chen14,chen15,lorbert12,sabuncu10} for more information. The responses of voxels in the ventral temporal cortex (VT) are collected to align the data while subjects watched the movie. Figure \ref{fig:Movies} illustrates the generated results for these two datasets. As depicted in this figure, we rank order the voxels by employing the voxel selection method in \cite{haxby11,chen14}; and the experiments are repeated by using the different number of ranked voxels, i.e. 100, 200, 400, 600, 800, 1000, and 1200. In addition, the empirical studies are reported by using the first 100 TRs, 400 TRs, 800 TRs, and 2000 TRs in both datasets. Figure \ref{fig:Movies} illustrates that the LDHA achieves superior performance to other HA algorithms. As mentioned before, our method can generate a better functional alignment of neural activities when the concept of the locality used for functional alignments. This improvement is more significant when the number of TRs in figure \ref{fig:Movies} is limited.
		
\section{Conclusion}
One of the main challenges in fMRI studies, especially MVP analysis, is using multi-subject datasets. On the one hand, the multi-subject analysis is necessary to estimate the validity of the generated results across subjects. On the other hand, analyzing multi-subject fMRI data requires accurate functional alignment between neuronal activities of different subjects for improving the performance of the final results. Hyperalignment (HA) is one of the most effective functional alignment methods, which can be formulated as a CCA problem for aligning neural activities of different subjects to a common space. The HA solution in MVP analysis may not be optimum because it mostly utilizes the unsupervised CCA techniques for functional alignment. This paper proposes the Local Discriminant Hyperalignment (LDHA) as a novel supervised HA solution, which employs the concept of locality in machine learning for improving the performances of both functional alignment and MVP analysis. Indeed, this paper defines the locality based on the stimuli categories (class labels) in the train-set. In a nutshell, the proposed method firstly generates two sets for each category of stimuli, i.e. the set of homogeneous stimuli as within-class neighborhoods and the set of stimuli from distinct categories as between-class neighborhoods. Then, these two sets are used to provide a better HA solution, where the correlation between the homogeneous stimuli is maximized, and also the correlation between different categories of stimuli is near to zero. Experimental studies on multi-subject MVP analysis demonstrate that the LDHA method achieves superior performance to other state-of-the-art HA algorithms. In the future, we will plan to develop a kernel-based version of LDHA for improving its performance in non-linear problems. 
\section*{Acknowledgment}
We thank the anonymous reviewers for comments. This work was supported in part by the National Natural Science Foundation of China (61422204 and 61473149), Jiangsu Natural Science Foundation for Distinguished Young Scholar (BK20130034) and NUAA Fundamental Research Funds (NE2013105).
\bibliographystyle{aaai17}
\bibliography{aaai17}

\begin{thebibliography}{}

\bibitem[\protect\citeauthoryear{Chen \bgroup et al\mbox.\egroup
  }{2014}]{chen14}
Chen, P.~H.; Guntupalli, J.~S.; Haxby, J.~V.; and Ramadge, P.~J.
\newblock 2014.
\newblock Joint svd-hyperalignment for multi-subject fmri data alignment.
\newblock In {\em IEEE International Workshop on Machine Learning for Signal
  Processing (MLSP)},  1--6.
\newblock IEEE.

\bibitem[\protect\citeauthoryear{Chen \bgroup et al\mbox.\egroup
  }{2015}]{chen15}
Chen, P. H.~C.; Chen, J.; Yeshurun, Y.; Hasson, U.; Haxby, J.; and Ramadge,
  P.~J.
\newblock 2015.
\newblock A reduced-dimension fmri shared response model.
\newblock In {\em Advances in Neural Information Processing Systems},
  460--468.

\bibitem[\protect\citeauthoryear{Conroy \bgroup et al\mbox.\egroup
  }{2009}]{conroy09}
Conroy, B.; Singer, B.; Haxby, J.; and Ramadge, P.~J.
\newblock 2009.
\newblock fmri-based inter-subject cortical alignment using functional
  connectivity.
\newblock In {\em Advances in Neural Information Processing Systems},
  378--386.

\bibitem[\protect\citeauthoryear{Dmochowski \bgroup et al\mbox.\egroup
  }{2012}]{dmochowski12}
Dmochowski, J.~P.; Sajda, P.; Dias, J.; and Parra, L.~C.
\newblock 2012.
\newblock Correlated components of ongoing eeg point to emotionally laden
  attention--a possible marker of engagement?
\newblock {\em Frontiers in Human Neuroscience} 6:112.

\bibitem[\protect\citeauthoryear{Duncan \bgroup et al\mbox.\egroup
  }{2009}]{duncan09}
Duncan, K.~J.; Pattamadilok, C.; Knierim, I.; and Devlin, J.~T.
\newblock 2009.
\newblock Consistency and variability in functional localisers.
\newblock {\em NeuroImage} 46(4):1018--1026.

\bibitem[\protect\citeauthoryear{Figueiredo and Nowak}{2016}]{figueiredo16}
Figueiredo, M.~A., and Nowak, R.~D.
\newblock 2016.
\newblock Ordered weighted l1 regularized regression with strongly correlated
  covariates: Theoretical aspects.
\newblock In {\em Proceedings of the 19th International Conference on
  Artificial Intelligence and Statistics},  930--938.

\bibitem[\protect\citeauthoryear{Gower and Dijksterhuis}{2004}]{gower04}
Gower, J.~C., and Dijksterhuis, G.~B.
\newblock 2004.
\newblock {\em Procrustes problems}, volume~30.
\newblock Oxford University Press on Demand.

\bibitem[\protect\citeauthoryear{Guntupalli \bgroup et al\mbox.\egroup
  }{2016}]{guntupalli16}
Guntupalli, J.~S.; Hanke, M.; Halchenko, Y.~O.; Connolly, A.~C.; Ramadge,
  P.~J.; and Haxby, J.~V.
\newblock 2016.
\newblock A model of representational spaces in human cortex.
\newblock {\em Cerebral Cortex}  bhw068.

\bibitem[\protect\citeauthoryear{Hanke \bgroup et al\mbox.\egroup
  }{2014}]{hanke14}
Hanke, M.; Baumgartner, F.~J.; Ibe, P.; Kaule, F.~R.; Pollmann, S.; Speck, O.;
  Zinke, W.; and Stadler, J.
\newblock 2014.
\newblock A high-resolution 7-tesla fmri dataset from complex natural
  stimulation with an audio movie.
\newblock {\em Scientific Data} 1.

\bibitem[\protect\citeauthoryear{Hardoon, Szedmak, and
  Shawe~Taylor}{2004}]{hardoon04}
Hardoon, D.~R.; Szedmak, S.; and Shawe~Taylor, J.
\newblock 2004.
\newblock Canonical correlation analysis: An overview with application to
  learning methods.
\newblock {\em Neural Computation} 16(12):2639--2664.

\bibitem[\protect\citeauthoryear{Haxby \bgroup et al\mbox.\egroup
  }{2011}]{haxby11}
Haxby, J.~V.; Guntupalli, J.~S.; Connolly, A.~C.; Halchenko, Y.~O.; Conroy,
  B.~R.; Gobbini, M.~I.; Hanke, M.; and Ramadge, P.~J.
\newblock 2011.
\newblock A common, high-dimensional model of the representational space in
  human ventral temporal cortex.
\newblock {\em Neuron} 72(2):404--416.

\bibitem[\protect\citeauthoryear{Haxby, Connolly, and
  Guntupalli}{2014}]{haxby14}
Haxby, J.~V.; Connolly, A.~C.; and Guntupalli, J.~S.
\newblock 2014.
\newblock Decoding neural representational spaces using multivariate pattern
  analysis.
\newblock {\em Annual Review of Neuroscience} 37:435--456.

\bibitem[\protect\citeauthoryear{Lorbert and Ramadge}{2012}]{lorbert12}
Lorbert, A., and Ramadge, P.~J.
\newblock 2012.
\newblock Kernel hyperalignment.
\newblock In {\em Advances in Neural Information Processing Systems},
  1790--1798.

\bibitem[\protect\citeauthoryear{Michael \bgroup et al\mbox.\egroup
  }{2015}]{michael15}
Michael, A.~M.; Anderson, M.; Miller, R.~L.; Adal{\i}, T.; and Calhoun, V.~D.
\newblock 2015.
\newblock Preserving subject variability in group fmri analysis: performance
  evaluation of gica vs. iva.
\newblock {\em Distributed Networks-New Outlooks on Cerebellar Function}  106.

\bibitem[\protect\citeauthoryear{Mohr \bgroup et al\mbox.\egroup
  }{2015}]{mohr15}
Mohr, H.; Wolfensteller, U.; Frimmel, S.; and Ruge, H.
\newblock 2015.
\newblock Sparse regularization techniques provide novel insights into outcome
  integration processes.
\newblock {\em NeuroImage} 104:163--176.

\bibitem[\protect\citeauthoryear{Oswal \bgroup et al\mbox.\egroup
  }{2016}]{oswal16}
Oswal, U.; Cox, C.; Lambon-Ralph, M.; Rogers, T.; and Nowak, R.
\newblock 2016.
\newblock Representational similarity learning with application to brain
  networks.
\newblock In {\em Proceedings of The 33rd International Conference on Machine
  Learning},  1041--1049.

\bibitem[\protect\citeauthoryear{Peng, Zhang, and Zhang}{2010}]{peng10}
Peng, Y.; Zhang, D.; and Zhang, J.
\newblock 2010.
\newblock A new canonical correlation analysis algorithm with local
  discrimination.
\newblock {\em Neural Processing Letters} 31(1):1--15.

\bibitem[\protect\citeauthoryear{Rademacher \bgroup et al\mbox.\egroup
  }{1993}]{rademacher93}
Rademacher, J.; Caviness, V.~S.; Steinmetz, H.; and Galaburda, A.
\newblock 1993.
\newblock Topographical variation of the human primary cortices: implications
  for neuroimaging, brain mapping, and neurobiology.
\newblock {\em Cerebral Cortex} 3(4):313--329.

\bibitem[\protect\citeauthoryear{Sabuncu \bgroup et al\mbox.\egroup
  }{2010}]{sabuncu10}
Sabuncu, M.~R.; Singer, B.~D.; Conroy, B.; Bryan, R.~E.; Ramadge, P.~J.; and
  Haxby, J.~V.
\newblock 2010.
\newblock Function-based intersubject alignment of human cortical anatomy.
\newblock {\em Cerebral Cortex} 20(1):130--140.

\bibitem[\protect\citeauthoryear{Sch{\"o}nemann}{1966}]{schonemann66}
Sch{\"o}nemann, P.~H.
\newblock 1966.
\newblock A generalized solution of the orthogonal procrustes problem.
\newblock {\em Psychometrika} 31(1):1--10.

\bibitem[\protect\citeauthoryear{Smola and Sch{\"o}lkopf}{2004}]{smola04}
Smola, A.~J., and Sch{\"o}lkopf, B.
\newblock 2004.
\newblock A tutorial on support vector regression.
\newblock {\em Statistics and Computing} 14(3):199--222.

\bibitem[\protect\citeauthoryear{Sui \bgroup et al\mbox.\egroup }{2011}]{sui11}
Sui, J.; Pearlson, G.; Caprihan, A.; Adali, T.; Kiehl, K.~A.; Liu, J.;
  Yamamoto, J.; and Calhoun, V.~D.
\newblock 2011.
\newblock Discriminating schizophrenia and bipolar disorder by fusing fmri and
  dti in a multimodal cca+ joint ica model.
\newblock {\em NeuroImage} 57(3):839--855.

\bibitem[\protect\citeauthoryear{Sui \bgroup et al\mbox.\egroup }{2013}]{sui13}
Sui, J.; He, H.; Pearlson, G.~D.; Adali, T.; Kiehl, K.~A.; Yu, Q.; Clark,
  V.~P.; Castro, E.; White, T.; Mueller, B.~A.; et~al.
\newblock 2013.
\newblock Three-way (n-way) fusion of brain imaging data based on mcca+ jica
  and its application to discriminating schizophrenia.
\newblock {\em NeuroImage} 66:119--132.

\bibitem[\protect\citeauthoryear{Talairach and Tournoux}{1988}]{talairach88}
Talairach, J., and Tournoux, P.
\newblock 1988.
\newblock {\em Co-planar stereotaxic atlas of the human brain. 3-Dimensional
  proportional system: an approach to cerebral imaging}.
\newblock Thieme.

\bibitem[\protect\citeauthoryear{Tom \bgroup et al\mbox.\egroup }{2007}]{tom07}
Tom, S.~M.; Fox, C.~R.; Trepel, C.; and Poldrack, R.~A.
\newblock 2007.
\newblock The neural basis of loss aversion in decision-making under risk.
\newblock {\em Science} 315(5811):515--518.

\bibitem[\protect\citeauthoryear{Wakeman and Henson}{2015}]{wakeman15}
Wakeman, D.~G., and Henson, R.~N.
\newblock 2015.
\newblock A multi-subject, multi-modal human neuroimaging dataset.
\newblock {\em Scientific Data} 2.

\bibitem[\protect\citeauthoryear{Watson \bgroup et al\mbox.\egroup
  }{1993}]{watson93}
Watson, J.~D.; Myers, R.; Frackowiak, R. S.~J.; Hajnal, J.~V.; Woods, R.~P.;
  Mazziotta, J.~C.; Shipp, S.; and Zeki, S.
\newblock 1993.
\newblock Area v5 of the human brain: evidence from a combined study using
  positron emission tomography and magnetic resonance imaging.
\newblock {\em Cerebral Cortex} 3(2):79--94.

\bibitem[\protect\citeauthoryear{Xu \bgroup et al\mbox.\egroup }{2012}]{xu12}
Xu, H.; Lorbert, A.; Ramadge, P.~J.; Guntupalli, J.~S.; and Haxby, J.~V.
\newblock 2012.
\newblock Regularized hyperalignment of multi-set fmri data.
\newblock In {\em IEEE Statistical Signal Processing Workshop (SSP)},
  229--232.
\newblock IEEE.

\end{thebibliography}
\end{document}